\documentclass[a4paper]{article}
\usepackage[left=4.4cm, right=4.4cm, top=5.2cm, bottom=5.2cm]{geometry} 
\usepackage[dvips]{graphicx}
\usepackage[latin1]{inputenc}
\usepackage{amssymb,amsmath,array}
\usepackage{enumitem}
\usepackage{multirow}
\usepackage{placeins}
\usepackage{xcolor,colortbl}
\usepackage{booktabs} 
\usepackage{adjustbox}
\setlist[itemize]{topsep=7pt, partopsep=7pt}

\begin{document}
\title{Mixed-Density Diffuser: Efficient Planning with Non-Uniform Temporal Resolution}

\author{Crimson Stambaugh and Rajesh P. N. Rao
%
\vspace{.3cm}\\
%
Paul G. Allen School of Computer Science and Engineering,\\University of Washington, Seattle, USA\\
}
\date{}
\maketitle
\thispagestyle{empty}

\begin{abstract}
 Recent studies demonstrate that diffusion planners benefit from sparse-step planning over single-step planning. Training models to skip steps in their trajectories helps capture long-term dependencies without additional memory or computational cost. However, predicting excessively sparse plans degrades performance. We hypothesize this temporal density threshold is  non-uniform across a planning horizon and that certain parts of a predicted trajectory should be more densely generated. We propose Mixed-Density Diffuser (MDD), a diffusion planner where the densities throughout the horizon are tunable hyperparameters. We show that MDD surpasses the SOTA Diffusion Veteran (DV) framework across the Maze2D, Franka Kitchen, and Antmaze Datasets for Deep Data-Driven Reinforcement Learning (D4RL) task domains, achieving a new SOTA on the D4RL benchmark.
\end{abstract}

\vspace*{-0.2in}
\section{Introduction}
Training a policy with online rollouts can be costly, dangerous, and sample-inefficient \cite{tutorial}. Alternatively, offline reinforcement learning (RL) involves a policy trained exclusively with pre-collected data.Extracting effective policies without exploration or feedback from the environment is challenging for conventional off-policy and even specialized offline RL algorithms \cite{iql}. Approaches to offline RL are also frequently faced with incomplete or undirected demonstrations \cite{wu2020behavior}. Offline algorithms must compose sub-trajectories from training data to generate advantageous behaviors. Other challenges are high-dimensionality and long horizons, which make accurate planning and behavior cloning difficult  \cite{tutorial}. Finally, sparse rewards pose a challenge to many training algorithms as they hinder accurate  credit assignment to actions \cite{d4rl}.

Diffusion models have emerged as a powerful framework for expressing complex, multi-modal distributions \cite{ddpm, song2021denoising}. Leveraging this model class, diffusion policies generate high fidelity actions and use a value function for action selection \cite{DQL,IDQL, SfBC}. Diffusion planners approach offline RL as a guided sequence generation task \cite{diffuser, DD, adaptdiffuser}. Planners predicting low temporal density, sparse-step trajectories are a promising approach to lengthening temporal horizons while maintaining manageable state-space output dimensionality \cite{HD}. However, excessively low density planning degrades performance, especially in tasks with critical short-horizon requirements, such as locomotion \cite{veteran}. This suggests a fundamental limitation of using uniform-density trajectories, which balance temporal resolution and detail with horizon length. We hypothesize that some parts of a planned trajectory require higher resolution than others \cite{lite}. Since uniform-density planners are unable to accommodate these shifts in redundant information, they often under-plan or model superfluous states.

Hierarchical planners circumvent this issue by training an ensemble of models with varying temporal resolutions and horizon lengths \cite{HD, HDMI}. In hierarchical approaches, a low-density planner creates coarse, long-horizon trajectories. Long horizon information, such as subgoals, is passed to a higher-density model which interpolates between the coarse states. Hierarchical models can achieve impressive performance with competitive inference and training times, but their ensemble structure means multiplying memory requirements and the number of trained parameters. The interdependence of a hierarchical ensemble hinders robustness as inaccuracies in the higher-level plan are compounded by errors in the lower-level model's predictions. Also, using a hierarchical framework with separate training of the levels makes end-to-end planner training impossible, potentially inhibiting optimization and information flow.

To overcome the limitations of both uniform-density and hierarchical planners, we introduce Mixed-Density Diffuser (MDD). Our main contributions are:\\
\textbf{1.} MDD generates trajectories with non-uniform temporal densities using a single, flat diffusion model. Our implementation integrates  mixed-density training with the powerful Diffusion Veteran (DV) framework \cite{veteran}.\\
\textbf{2.} MDD achieves state-of-the-art performance on several of the Datasets for Deep Data-Driven Reinforcement Learning (D4RL)  tasks \cite{d4rl}, surpassing the performance of the previous SOTA DV framework, without additional model weights or inference computation costs. \\
\textbf{3.} We present strong empirical evidence supporting our hypothesis that non-uniform temporal horizons are a key principle of efficient and effective diffusion planning.
\vspace*{-0.2in}
\begin{figure}[!htb]
  \centering
  \includegraphics[width=0.65\linewidth]{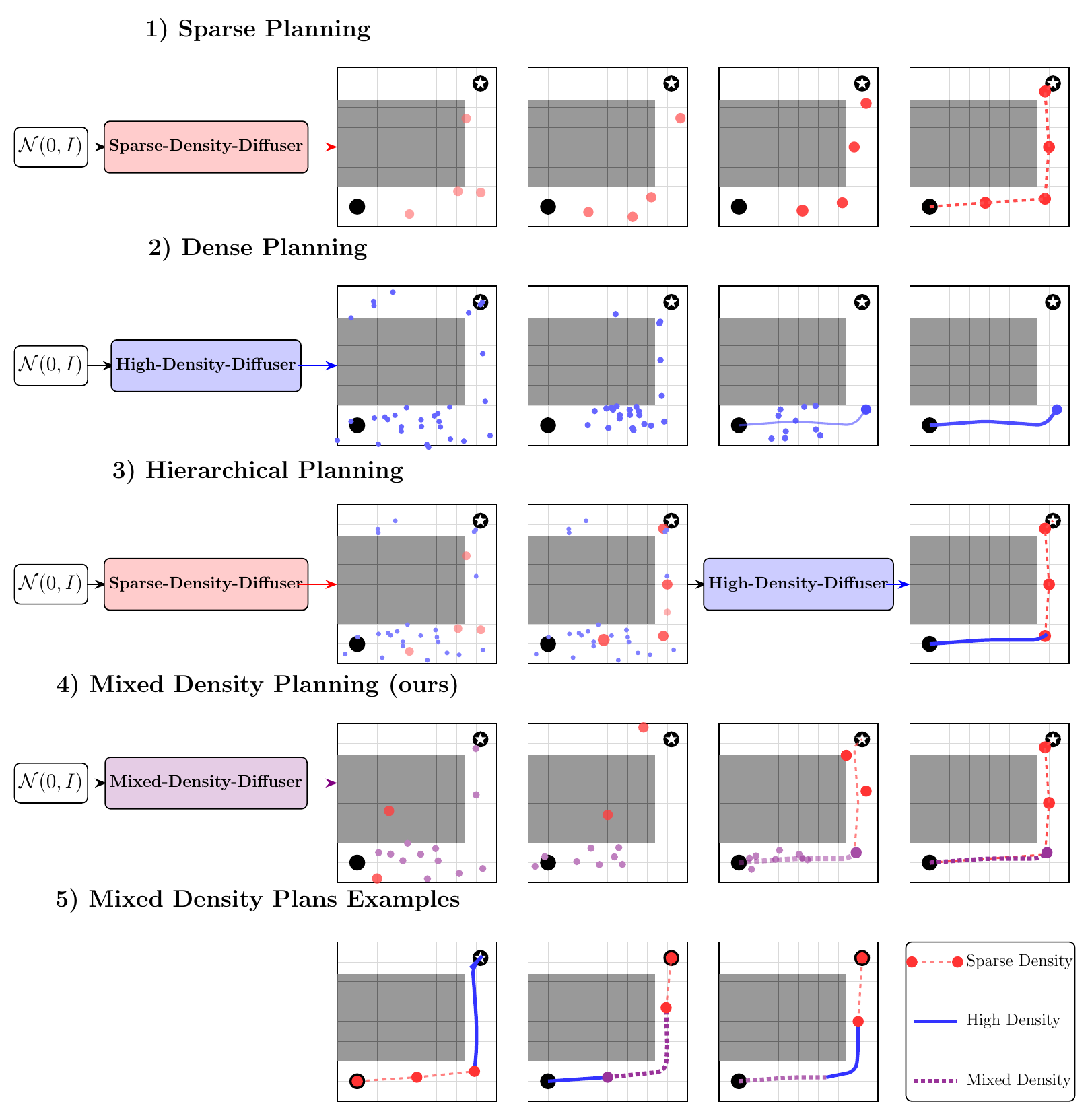}
  \caption{\footnotesize{{\bf Comparison of trajectory denoising in different diffusion planning approaches}. Sparse Density Diffusers in row $1)$  extend temporal planning horizons with comparatively little computational cost at the price of low temporal resolution. High Density Diffusers in row $2)$ compute many steps for shorter temporal horizons creating more continuous planned trajectories. Hierarchical Planners in row $3)$ \cite{HD,lite, HDMI} generate way-points with a Sparse Density Diffuser and interpolate between with a High-Density Diffuser. MDD (bottom row) utilizes the benefits of sparse and dense planning with a simple, flat framework by treating the sparse and dense observations as a single trajectory for one planner to generate. Row $5)$ shows fully denoised example trajectories that would be non-trivial to generate with Hierarchical Planners, but simple to implement with MDD.}}
  \label{fig:logo}
\end{figure}
\section{Diffusion Planning}
Diffusion planners \cite{diffuser,DD,adaptdiffuser, veteran} approach offline RL as a guided sampling problem. Given a state $s_t$ or state-action pair $(s_t, a_t)$ at time $t$, diffusion planners predict an $H$-step trajectory, $\boldsymbol{\tau}$, where the jump variable $K$ represents the interval between planned timesteps. The initial position $X_t$ is fixed to the current true observation. A trajectory output where $X_t$ can either be $s_t$ or  $(s_t, a_t)$ is:
\[
\boldsymbol{\tau}= \begin{bmatrix} X_t & X_{t+K} & X_{t+2K} & \dots & X_{t+HK} \end{bmatrix}
\]A guidance function models the expected value of trajectories and focuses sampling toward high-reward trajectories. Early implementations predicted state-action trajectories, but recent studies \cite{DD,veteran} empirically show generating state-only trajectories and using an inverse dynamics model $h$ yields higher performance, where $h(s_t,s_{t+k}) = a_t$, an action that is likely to move the agent to state $s_{t+k}$ from $s_t$. Unlike auto-regressive world models traditionally used in model-based RL \cite{planet}, these diffusion planners predict states at the trajectory level, avoiding compounding errors from one-step rollouts using world models. In existing diffusion planners, the jump variable $K$ is constant, preventing precise temporal resolution allocation--a limitation our MDD planner addresses.
\vspace*{-0.2in}
\section{Mixed-Density Diffuser Framework}
While our approach can be flexibly integrated with many diffusion planner designs, we build on the DV framework, which is the result of a comprehensive hyperparameter sweep \cite{veteran}. MDD employs a Diffusion Transformer (DiT) denoising backbone \cite{Peebles2022DiT}, Monte Carlo Sampling guidance, and a single flat planner that predicts state-only trajectories, paired with a diffusion-based inverse dynamics model. 

We can abstract the output trajectory as:
\[
\boldsymbol{\tau} = 
\begin{bmatrix}
x_t & x_{t+K_1} & x_{t+K_1+K_2} & \dots & x_{t+K_1+K_2+\dots+K_{H-1}}
\end{bmatrix}
\]
where each $K_i$ is an independent, tunable hyperparameter (in the current implementation, these are manually tuned). Adjusting these $K_i$ jump sizes allows precise control over temporal resolution. For example, our model for Franka Kitchen (see below) denoises an output trajectory of $H=32$ entries, where $K_i=4$ for $i=1$ through $10$ and $K_i=6$ for $i=11$ through $31$. Manually tuning ranges of $K_i$'s (rather than each individual $K_i$) quickly narrowed down our hyperparameter search and improved interpretability. We leave the question of learning optimal $K_i$ ranges and values to future work.

\vspace*{-0.2in}
\section{Experimental Results}
We evaluate MDD on three offline RL task domains from the widely used D4RL benchmark~\cite{d4rl}: Maze2D, AntMaze, and Franka Kitchen \cite{lynch2019play}. In \textbf{Maze2D}, the agent must navigate a 2D maze using a simple ball embodiment, directly testing MDD's long-horizon planning ability. \textbf{Antmaze} extends Maze2D to a higher-dimensional setting by replacing the ball with an 8-DoF quadruped. This is a sparse rewards problem as the agent receives no rewards except a reward of 1 when it reaches the target location. In \textbf{Franka Kitchen}, the agent is trained using suboptimal or incomplete demonstrations to control a 9-DoF Franka arm for a multi-part manipulation task. We compare MDD against several planner-based and policy-only baselines. For these baselines, we report the results from their original publications to ensure they are represented with tuned hyperparameters for D4RL. Our main results are displayed in Table ~\ref{tab:performance}. 
\vspace*{-0.2in}
\begin{table}[ht]
\caption{Normalized score comparison of offline RL methods across 150 episode seeds on the Franka Kitchen \cite{lynch2019play}, Antmaze, and Maze2D D4RL tasks \cite{d4rl}. The best scores per column are bolded.}
\label{tab:performance}
\adjustbox{max width=\textwidth}{
\begin{tabular}{l|cc>{\columncolor{gray!20}}c|cccc>{\columncolor{gray!20}}c|ccc>{\columncolor{gray!20}}c}\toprule
& \multicolumn{3}{c|}{\textit{Kitchen}} & \multicolumn{5}{c|}{\textit{Antmaze}} & \multicolumn{4}{c}{\textit{Maze2D}} \\
\cmidrule(lr){2-4} \cmidrule(lr){5-9} \cmidrule(lr){10-13}
\textit{Method} & \textit{Mixed} & \textit{Partial} & \textit{avg.} & \textit{L.-diverse} & \textit{L.-play} & \textit{M.-diverse} & \textit{M.-play} & \textit{avg.} & \textit{L} & \textit{M.} & \textit{Umaze} & \textit{avg.}\\ \midrule
BC & 47.5 & 33.8 & 40.7 & 0.0 & 0.0 & 0.0 & 0.0 & 0.0 & 5 & 30.3 & 3.8 & 13.0 \\ \midrule
SfBC \cite{SfBC} & 45.4 & 47.9 & 46.7 & 45.5 & 59.3 & 82.0 & 81.3 & 67.0 & 74.4 & 73.8 & 73.9 & 74.0\\
DQL \cite{DQL} & 62.6 & 60.5 & 61.6 & 56.6 & 46.4 & 78.6 & 76.6 & 64.6 & - & - & - & - \\
DQL* \cite{cleandiffuser}& 55.1 & 65.5 & 60.3 & 70.6 & \textbf{81.3} & 82.6 & 87.3 & 80.5 & 186.8 & 152.0 & 140.6 & 159.8\\
IDQL \cite{IDQL} & 66.5 & 66.7 & 66.6 & 67.9 & 63.5 & 84.8 & 84.5 & 75.2 & 90.1 & 89.5 & 57.9 & 79.2 \\
IDQL* \cite{cleandiffuser}& 66.5 & 66.7 & 66.6 & 40.0 & 48.7 & 83.3 & 67.3 & 69.8 & - & - & - & - \\ \midrule
Diffuser \cite{diffuser}& 52.5 & 55.7 & 54.1 & 27.3 & 17.3 & 2.0 & 6.7 & 13.3 & 123 & 121.5 & 113.9 & 119.5 \\
DD \cite{DD} & \textbf{75.0} & 56.5 & 65.8 & 0.0 & 0.0 & 4.0 & 8.0 & 3.0 & - & - & - & - \\
HD \cite{HD} & 71.7 & 73.3 & 72.5 & \textbf{83.6} & - & 88.7 & - & - & 128.4 & 135.6 & \textbf{155.8} & 139.9 \\
DfsrLite \cite{lite} & 73.6 & 74.4 & 74.0 & 80.4 & 72.4 & \textbf{89.2} & 88.0 & 82.5 & - & - & - & -\\ \midrule
DV \cite{veteran} & 73.6 & 94.0 & 83.8 & 80.0 & 76.4 & 87.4 & \textbf{89.0} & 83.2 & 203.6 & 150.7 & 136.6 & 163.6  \\ \midrule 
MDD (Ours) & \textbf{75.0} & \textbf{99.7} & \textbf{87.4} & 80.7 & 77.3 & 87.3 & 88.7 & \textbf{83.5} & \textbf{206.1} & \textbf{154.5} & 138.3 & \textbf{166.3} \\ \bottomrule
\end{tabular}
}
\end{table}

\vspace*{-0.2in}
\section{Discussion}
 Our results show that by simply adjusting the temporal density of the output trajectory states, we can boost the performance of a diffusion planner on a variety of offline RL tasks. 
 
Following the DV publications' procedure \cite{veteran}, we used the same temporal density configuration for each task in a task domain to avoid brittle per-task tuning. Note that DV's hyperparameters, including its temporal planning density, were selected from a hyperparameter sweep of over $6,000$ models \cite{veteran}. Thus, DV is a strong baseline model for MDD. We used a denser-to-sparser density configuration but other density configurations may yield superior performance on specific subtasks. For example, a sparse-to-dense configuration actually surpassed every model we compared in the Antmaze-Diverse task. We hypothesize that in this case, increasing the temporal resolution in the latter part of the output horizon provided finer-grained supervision for dependencies spanning long time horizons. We leave further investigation of this result for future work. 
 
 Although MDD is outperformed slightly by one or two models on five subtasks in Table ~\ref{tab:performance}, the base DV framework was outperformed by an even larger margin on half of MDD's underperforming subtasks. This suggests a single mixed-density configuration does not always generalize well to every subtask and that MDD inherits some of DV's shortcomings. In particular, MDD and DV achieved very high performance on the Large and Medium Maze2D tasks, but were surpassed by two baselines on the Umaze task. The Umaze environment is smaller and simpler than the Large and Medium environments suggesting that approaches like HD and DQL have superior short-term planning but lack the long-horizon foresight to achieve a higher mean score on the Maze2D tasks. In both the Kitchen and Maze2D tasks, we compare two variations of the Diffusion Veteran framework to test whether MDD's performance increase came from simply increasing the sparsity and temporal horizon of generated DV trajectories. We adopt the notation DV-J where J is the adjusted jump size. Surprisingly, we find that increasing DV's jump size actually significantly increases its Kitchen performance despite DV having already been extensively tuned. MDD still outperforms these DV-J variants on average.

 \vspace*{-0.2in}
\begin{table}[ht]
	\caption{DV temporal density variations of the Franka Kitchen and Maze2D tasks. Results are averaged over 1000 episode seeds}
	\label{tab:locations}
	\adjustbox{max width=\textwidth}{
    \begin{tabular}{rll>{\columncolor{gray!20}}lllll>{\columncolor{gray!20}}l>{\columncolor{gray!20}}l}\toprule
		\textit{Method} & \textit{Kitchen Mixed} & \textit{Kitchen Partial} & \textit{avg.} & \textit{Method} & \textit{Maze L.} & \textit{Maze M.} & \textit{Umaze} & \textit{avg.}\ \\ \midrule
		DV-5  & 74.7 & 99.4 & 87.05 & DV-17 & 125.9 & 150.8 & 195.3 & 157.33\\
		DV-6  & 74.65 & 98.5 & 86.6 & DV-25 & 130.4 & \textbf{153.5} & \textbf{199.5}& 161.13\\
        MDD & \textbf{75.0} & \textbf{99.7} & \textbf{87.4} & MDD & \textbf{206.1} & 151.5 & 138.2 & \textbf{166.3}\\
        \bottomrule
        \end{tabular}
    }
\end{table}
 \vspace*{-0.2in}
\begin{footnotesize}
\bibliographystyle{unsrt}
\bibliography{references}
\end{footnotesize}

\end{document}